\documentclass[conference]{IEEEtran}
\usepackage{cite}
\usepackage{amsmath,amssymb,amsfonts}
\usepackage{algorithmic}
\usepackage{graphicx}
\usepackage{textcomp}
\usepackage{pdfpages}
\usepackage{xcolor}

\usepackage{xspace}
\usepackage{hyperref}
\usepackage[colorinlistoftodos]{todonotes}
\usepackage{verbatim}

\begin{document}

%%%%%%%%%%%%%%%

\newcommand\acronym{\textit{Savvy}\xspace}
\newcommand\ps{\mathcal{P}\xspace}
\newcommand\cs{\mathcal{C}\xspace}

\title{\acronym: Trustworthy Autonomous Vehicles Architecture \\ \small{(PREPRINT)}}

\author{\IEEEauthorblockN{Ali Shoker, Rehana Yasmin, and Paulo Esteves-Verissimo
}
\IEEEauthorblockA{\textit{Resilient Computing and Cybersecurity Center (RC3),} \\
\textit{Computer, Electrical and Mathematical Sciences and Engineering Division (CEMSE),} \\
\textit{King Abdullah University of Science and Technology (KAUST)}\\
Thuwal 23955-6900, Kingdom of Saudi Arabia \\
\{ali.shoker, rehana.yasmin, paulo.verissimo\}@kaust.edu.sa}
}

% \IEEEoverridecommandlockouts
% \makeatletter\def\@IEEEpubidpullup{6.5\baselineskip}\makeatother
% \IEEEpubid{\parbox{\columnwidth}{
%     Symposium on Vehicles Security and Privacy (VehicleSec) 2023 \\
%     27 February 2023, San Diego, CA, USA \\
%     ISBN 1-891562-88-6 \\
%     https://dx.doi.org/10.14722/vehiclesec.2023.23xxx \\
%     www.ndss-symposium.org
% }
% \hspace{\columnsep}\makebox[\columnwidth]{}}

\maketitle

%%%%%%%%%%%%%%%%%%%

\begin{abstract}
The increasing interest in Autonomous Vehicles (AV) is notable due to business, safety, and performance reasons. While there is salient success in recent AV architectures, hinging on the advancements in AI models, there is a growing number of fatal incidents that impedes full AVs from going mainstream. This calls for the need to revisit the fundamentals of building safety-critical AV architectures. However, this direction should not deter leveraging the power of AI. To this end, we propose \acronym, a new trustworthy intelligent AV architecture that achieves the best of both worlds. \acronym makes a clear separation between the \textit{control plane }and the \textit{data plane} to guarantee the \textit{safety-first principles}. The former assume control to ensure safety using design-time defined rules, while launching the latter for optimising decisions \textit{as much as possible} within safety time-bounds. This is achieved through guided \textit{Time-aware predictive quality degradation (TPQD)}: using dynamic ML models that can be tuned to provide \textit{either} richer \textit{or} faster outputs based on the available safety time bounds. For instance, \acronym allows to safely identify an elephant as an obstacle (a mere object) the earliest possible, rather than optimally recognizing it as an elephant when it is too late. This position paper presents the \acronym's motivations and concept, whereas empirical evaluation is a work in progress.

\end{abstract}

% No keywords for VehicleSec
\begin{IEEEkeywords}
Autonomous Driving, Autonomous Vehicle, ADAS, Safety, Architecture
\end{IEEEkeywords}
\section{Introduction}
\label{sec:intro}

There is an unprecedented interest in building Autonomous Vehicles (AV) by leveraging Artificial Intelligence (AI), motivated by business, safety, and performance reasons. 
The AV market is in steady growth\footnote{https://www.bloomberg.com/press-releases/2023-02-03/autonomous-cars-market-is-expected-to-generate-a-revenue-of-usd-15-55-billion-by-2030-globally-at-31-19-cagr-verified-market} at a CAGR of 30\%, hitting the market size\footnote{https://www.statista.com/statistics/1224515/av-market-size-worldwide-forecast/} of 2 Trillion USD by 2030. %, and generating 
%20 Billion USD revenue by 2030, and 
%400 Billion USD by 2035
%\footnote{https://www.mckinsey.com/industries/automotive-and-assembly/our-insights/autonomous-drivings-future-convenient-and-connected}.
While the performance motivations of AVs are obvious, e.g., through optimizing the driving experience through situational awareness, the safety benefits are little understood. 
A recent study by RAND~\cite{kalra2017enemy-AV-RAND} argues that "delaying full deployment of AVs until an extraordinarily 18 high level of safety is achieved in comparison to human drivers could cost hundreds of thousands of lives over many years"~\cite{webb2020waymo}. One can infer that, regrettably, the success of AVs is on the premise of building safe solutions that shall not kill humans more than conventional, non-autonomous, vehicles typically do. Nevertheless, the increasing global fatal incidents of AVs do not look satisfactory or encouraging~\cite{driver-deaths-incidents,tesla-deaths-incidents,shah2019safe-AV-Simplex-EGAS}, urging on the need to revisit the fundamentals of building safety-critical AV architectures and solutions.

There is a trade-off between safety and efficiency properties in AV architectures. While we adopt the "Safety First for Automated Driving" (SaFAD)~\cite{CLEPA-safety-first-degrad-2019} in this work, which is also promoted by the European Association of Automotive Suppliers, a practical AV architecture must account for performance. The challenge is to build AV architectures that can find the right balance.

Current AV architectures capitalize heavily on the recent advances in AI/ML. Unfortunately, AV architectures in production like \textit{Tesla, Cruise, Waymo}, and \textit{Udacity} have gained bad reputation given the increasing death incidents~\cite{cruise-incidents,tesla-deaths-incidents,driver-deaths-incidents} attributed to the failure of AI/ML systems in particular. Our study to reported investigations, studied in Section~\ref{sec:motiv} concludes that the majority of these incidents are caused by the tendency to prioritize performance at the cost of safety. %This leads to the misconception that AI/ML is not adequate to AVs as long as accuracy is below seven nines accuracy. This is arguable since considering the main phases of IAV processing, i.e., recognition, planning, and actuation, AI/ML algorithms can be much more effective than human recognition and reaction. Indeed, such algorithms can leverage a set of sensors to make better situational awareness, and can better predict the environmental movements (e.g., multiple vehicles) to infer safer manoeuvring than humans. 
This has been raising concerns in the research community for a decade now, which motivated several academic works like Safe-AV~\cite{shah2019safe-AV-Simplex-EGAS}, Sentinel~\cite{deevy2019sentinel}, and \textit{KARYON}~\cite{casimiro2013karyon} to address this issue. These works have built on good early architectures and concepts, e.g., \textit{Simplex} architecture~\cite{2001-AV-Arch-Simplex,supervisor-simplex-redundancy-AV-2018} and \textit{E-GAS}~\cite{AV-ARch-safety-use-eGas}, that emphasized safety-first through building redundant systems that activate a \textit{single property at a time}: safety or performance. For instance the famous \textit{Simplex} architecture~\cite{2001-AV-Arch-Simplex,supervisor-simplex-redundancy-AV-2018} uses a control circuit to switch from a default efficient mode to a safe mode under faulty or hazardous situations. 
%While the concept is sound and prominently used in safety systems, its implementation can have serious flaws since under some failures the system running in the efficient mode may fail to handover to the safey mode~\cite{casimiro2013karyon}. This has been solved in two ways in literature. 
E-GAS~\cite{AV-ARch-safety-use-eGas} extends this with multi-layered supervisory layer monitors; whereas KARYON~\cite{casimiro2013karyon} used hardware-hardened safety kernel that takes-over instead of having a possibly failing system mode handover---that might never occur due to AI inference delays.
%While these approaches have also supported the safe operation with degradation, they do not exploit the advances on AI/ML for autonomous vehicles, which proves (as we discuss next) to be advantages in several situations when used right.
The dilemma is that these proposals tend to be too conservative since they activate the safe-operational control when the AI inference fails to deliver on time. This ends up giving up the power of AI to improve both performance and safety. The raised question is again: what is the right balance between safety and performance?

We conducted an analysis to real-world AV reported incidents~\cite{shah2019safe-AV-Simplex-EGAS,Tesla-S-real-issues-2017autonomous,SOTIF-Schwalb2019AnalysisOS} to understand the current phenomenona behind AV failures. Our conclusions refer these failures to two main reasons (discussed in Section~\ref{sec:motiv}): (1) Lost Command and Control and (2) AI-based AVs are optimized to deliver All-or-Nothing. The two reasons are very related: the system is often confused and cannot make a decision because the AI system has not delivered early enough, or never, before the incident. It seems that the AI system often tries to optimize detection, recognition, and planning which often exceeds the available time bounds. In many cases, one could also infer that these AVs  are using some safety-first concepts as those introduced in theory~\cite{2001-AV-Arch-Simplex,AV-ARch-safety-use-eGas,shah2019safe-AV-Simplex-EGAS}. Nevertheless, in the best case when it succeeds to fail-operational, the AV completely sacrifices the power and cost of the AI system.

\begin{table*}[ht]
    \centering
    \caption{A sample of investigated vehicle incidents with different levels of autonomy, referred to common potential causes.}
    \label{tab:incidents}
    \begin{tabular}{p{0.02\linewidth} p{0.15\linewidth} p{0.55\linewidth}  p{0.2\linewidth}}
    \hline
            \textbf{ID} & \hspace{2cm} \textbf{Incident} & \textbf{Description} & \textbf{Potential Causes}\\
    \hline\hline
        I1 & Uber Volvo XC90 (Arizona, 2018)~\cite{volvo-XC90-incidents,shah2019safe-AV-Simplex-EGAS} & A modified Volvo XC90 struck and killed a pedestrian walking a bike and crossing a road at night. The vehicle was equipped with a LIDAR unit, forward facing and side facing cameras, radars and Uber’s developmental AV software. The car did not brake or attempt to slow down to avoid the collision. Emergency breaking and driver monitoring were disabled. NTSB report states that the vehicle’s sensors detected the pedestrian 6 seconds before the accident; initially detecting her as an unknown object, then as a vehicle, and finally as a bicycle; 1.3 seconds before impact, the vehicle’s software determined that a braking action was required. &  (1) Emergency breaking and driver monitoring were disabled. (2) Braking decision has been made 4.7 seconds after first detection.
        \\\hline
        I2 & Acura MDX (New-foundland, 2018)~\cite{shah2019safe-AV-Simplex-EGAS} & Acura MDX equipped with Acura’s lane-keep assistance attempt to veer off of its lane and sometimes into oncoming traffic. This was noticed after replacing vehicle’s windshield. & The camera, a crucial component for lane-keep feature, was not calibrated after windshield replacement.
        \\\hline
        I3 & Tesla Model X (California, 2017)~\cite{shah2019safe-AV-Simplex-EGAS} & Tesla Model X (without LIDAR) while using Autopilot feature (cruise control and autosteer lane-keep) crashed into a damaged crash highway attenuator and fatally wounded its driver. At three seconds prior to the crash and up to the time of impact with the crash attenuator, the Tesla’s speed increased, with no precrash braking or evasive steering movement detected. & (1) Attenuator has not been detected yet. (2) No safety short circuit caused the car to slow-down or stop. (3) Sensors are not used effectively.
        \\\hline
        I4 & Tesla Model S (California, 2018)~\cite{shah2019safe-AV-Simplex-EGAS} & Tesla Model S (with Autopilot enabled) was travelling on a freeway crashed into a stopped fire truck. The Tesla was following another vehicle that swerved out of the lane to avoid the stopped fire truck, while the Tesla sped up instead, and crashed into the truck. & (1) Cruise Control failed to detect the stationary truck on time. (2) Cruise Control detected but ignored the truck.
        \\\hline
        I5 & Tesla Model S (Florida, 2016)~\cite{shah2019safe-AV-Simplex-EGAS} & Tesla Model S with Autopilot engaged struck and passed beneath a coming tractor trailer that was making a left turn infront of the Tesla from the westbound lanes of the highway across the two eastbound travel lanes. NTSB reported that the Tesla’s automated vehicle control system did not identify the truck crossing the car’s path or recognize the impending crash; consequently, the Autopilot system did not reduce the car’s velocity, the forward collision warning system did not provide an alert, and the automatic emergency braking did not activate. Tesla commented that the camera failed to detect the truck due to "white colour against a brightly lit sky" and a "high ride height", and that the radar filtered out the truck as an overhead road sign to prevent false braking. & (1) Truck crossing has not been detected on time. (2) Safety circuit has not engaged the emergency braking. (3) Sensors are not utilized effectively. (4) Detection has been ignored.
        \\\hline
        I6 & Tesla Model S (China, 2016)~\cite{shah2019safe-AV-Simplex-EGAS} & Tesla Model S crashed into a slow moving (or parked) street sweeper and killed its driver. The police concluded that the neither the driver nor the vehicle had attempted any braking or collision avoidance manoeuvres. Tesla was equipped with a single forward facing radar, a single forward facing camera and a set of 12 ultrasonic sensors. While the camera used DNN recognition models over MobileEye’s EyeQ3 computing platform, the system required agreement between both the camera and the radar before any action was taken. & (1) Camera system failed to detect the sweeper on time; (2) Camera and Radar both failed. (3) Detection of Radar alone has been ignored.  
        \\\hline
        I7 &  GM Cruise (San Francisco, 2022)~\cite{cruise-2022-incidents} & Cruise vehicle operating in autonomous mode made a left turn in front of an oncoming Toyota Prius and preformed hard brake at an intersection. NHTSA reported that the Cruise's ADAS could make "unprotected left, cause ADAS to incorrectly predict another vehicle’s path or be insufficiently reactive to the sudden path change of a road user." Cruise said the software had to decide between two different risk scenarios: hard break or collide before the oncoming vehicle’s sudden change of direction".  & (1) ADAS cannot predict path on time. (2) Pre-defined decisions (turn left) is not always reasonable.
        \\\hline

    \end{tabular}
\end{table*}

The above observations inspired us to explore guided \textit{Time-aware predictive quality degradation (TPQD)} that tunes AI models to enforce different richness levels or predictive quality dictated by the available safety time bounds. This allows to leverage the \textit{best outcome an AI system can deliver within a given time interval}. We show in this paper that, in many scenarios, AI models seem to be over-optimized to give rich predictive details, while basic details can be good-enough for situations where time is paramount. 

Based on these concepts, we propose \acronym, a new preliminary AV architecture that stands as a sweet spot between performance and safety. \acronym ensures the safety-first principle by combining safe-operational control together with TPQD. Ths is possible using Dynamic AI models that can be tuned to deliver before the safety-critical time expires. We are exploring Dynamic Neural Networks that allow for model deformation using depth and width adjustment~\cite{2016branchynet-dynamic-depth-exiting-NN,huang2017multi-dynamic-depth-exiting-NN,wang2018skipnet-CNN-skipping,lin2017runtime-width-prunning-NN,cai2021dynamic-width-skipping-branches-NN} (early exiting, skipping, pruning, etc.), choosing the adequate protocol using Neural Architecture Search~\cite{zoph2016neural-NAS-survey, yu2020bignas-BigNAS} (NAS), or parameter (Weights, Space, or Channel) adjustments~\cite{woo2018cbam-attention,guo2022attention-survey,wang2017deep-attention,chen2020dynamic-attention,li2021dynamic-slimmable-NN} at inference time. We are currently implementing a Proof of Concept of \acronym and driving an empirical evaluation of TPQD.

The rest of the paper is organized as follows. Section~\ref{sec:motiv} presents the motivations of \acronym through analyzing AV incident investigations. Section~\ref{sec:arch} introduces \acronym architecture, while Section~\ref{sec:related} discusses the related works. The paper concludes in Section~\ref{sec:conc}.

\section{The Case for \acronym}
%\todo[size=\small, color=green!40]{Rehana, plz review}%
\label{sec:motiv}

This work is driven by our concerns about the hundreds of vehicle incidents~\cite{shah2019safe-AV-Simplex-EGAS,Tesla-S-real-issues-2017autonomous,SOTIF-Schwalb2019AnalysisOS} related to autonomous driving features at any SAE level, including driving assistance, emergency brake, auto-steering, cruise controls, etc. Analyzing dozens of these reported incidents (mostly under investigations), we observed that despite their different circumstances, the majority can be attributed to a common set of potential causes conveyed in Table~\ref{tab:incidents}, which stands as a briefing of seven credible well-reported and investigated incidents. We inspire from these observations to explain the design rationale behind \acronym (discussed in Section`\ref{sec:arch}):

\subsection{Lost Command and Control}
We observed that many incidents occur because of the confusion in "command and control" in the autonomous driving (AD) system. In many cases, the control is either retained by the data plane, i.e., ML-based system, or "lost" because of some confusion in the system or handover. While an ML-based system does have promising potential, even for safety features, it cannot fully retain AD control because (1) it is a probabilistic solution, and (2) it does not oversee the entire vehicle state. For instance, incidents I3 through I7 convey different AD failures in detection or prediction. Worse, in some cases like I1, I3, and I5, the vehicle ignored the sensing alerts and has not made any slow-down or braking reactions; either because the AD system could not make that decision or because a last moment handover was being done. The AD system should have at least handed over the control to some safety-circuit in such a hazardous situation. On the other hand, many incidents like I1, I2, I4, and I6 are referred to the lack of overseeing or handling the vehicle posture, e.g., features disabled, broken or non-calibrated sensors/actuators, or radars sensing ignored. The vehicle state should be retained by a reliable system that can oversee the entire vehicle state as well as \textit{take over, not get handed over} the control in such critical situations where ML fails to respond on time.

\subsection{AI is optimized to deliver All-or-Nothing}
We noticed that in all incidents of Table~\ref{tab:incidents}, the ML-based AD failed to recognize an obstacle or predict a plan or a maneuver on time. The term "on time" here is key since no reports mentioned the AD returning an "invalid" or "indeterminate" classification or prediction, but rather AD \textit{has not delivered early enough before} the incident. Notice that in I1, for example, the AD system recognized an unknown object 6 seconds before the incident and took 4.7 seconds to determine a brake is required. I3 and I7 show that an obstacle has been detected, but the system was not able to make a correct prediction or decision on time. Incidents I3, I5, and I6 show that some sensors have not been considered in decision making, maybe for some optimization (avoiding false-negatives) or because the system could not make a timely sensor fusion. 

Our hypothesis is that \textit{ML-based solutions are over optimized to deliver All-or-Nothing regardless of the delivery time}. While this can be questionable, it inspired us to study the effectiveness of accepting some \textit{predictive quality degradation in ML inference in favor of timeliness} to guarantee safety. This helps in enforcing decision making timeouts, e.g., by calibrating dynamic ML algorithms~\cite{2016branchynet-dynamic-depth-exiting-NN,huang2017multi-dynamic-depth-exiting-NN,wang2018skipnet-CNN-skipping,lin2017runtime-width-prunning-NN,cai2021dynamic-width-skipping-branches-NN} (e.g., deepness or parameters) to deliver before timeouts as long as the output is helpful, though not optimal. 

To exemplify, we convey different scenarios presented in Table~\ref{tab:scenarios}. The table demonstrates that from a single event, there could be different rich/poor sensing and planning levels upon which "possible actions" can be made the earliest possible. In the Obstacle Avoidance scenario for instance, there are many sensing levels, either because of ML model deepness or considered sensors fusion, that could help making constructive decisions within some known time windows. Depending on the time availability (e.g., before hitting an obstacle), a used ML model is calibrated (or another model is fetched) to give richer details. Note that the possible actions in the table are not our recommendations, but rather used to explain the concept. (This is worth another study out of the scope of this paper.) This motivated us to designing \acronym, a new architecture with this concept as we show next.

\begin{table}[h]
    \centering
    \caption{Different scenarios showing that some time-aware predictive quality degradation can still be helpful in decision making within different time bounds. (Hint: Increasing L refers to richer ML output details, incurring more delays.)}
    \label{tab:scenarios}
    \begin{tabular}{p{0.05\linewidth} p{0.5\linewidth}  p{0.3\linewidth}}
    \hline
            \textbf{Level} & \hspace{2cm} \textbf{Sensing} & \textbf{Possible Action}\\
    \hline\hline\\
    &\hspace{1.5cm}         \textbf{ \textsc{1. Obstacle avoidance}}&\\

        L1 & An object detected at safety distance & breaks; beep \\\hline
        L2 & Non obstructive shaped (flat, small, short) object detected & continue\\\hline
        L3 & Non obstructive material object detected (rubber, herb plant, snow) & continue slowly\\\hline
        L4 & Obstructive avoidable object detected & beep; steer away\\\hline
        L5 & Obstructive unavoidable material object detected & breaks; beep\\\hline
        L6 & Obstructive mobile object detected (auto, animal) & breaks; give way; continue later\\\hline
        L7 & Obstructive rational object (human) detected & breaks; stop; continue later\\\\
          \hline\\
    &\hspace{1.2cm}         \textbf{\textsc{2. Intersection crossing}}&\\

        L1 & No cooperative sensing & breaks \\\hline
        L2 & Cooperative sensing (e.g, RSU) short distance & break\\\hline
        L3 & Cooperative sensing (e.g, RSU) long distance & continue\\\hline
        L4 & Cooperative active sensing & agreement \\
        \hline\\
    &\hspace{2.5cm}         \textbf{\textsc{3. Overtaking}}&\\
        L1 & No cooperative sensing & continue \\\hline
        L2 & Cooperative sensing (e.g, RSU) short distance & slow down\\\hline
        L3 & Cooperative sensing (e.g, RSU) long distance & overtake\\\hline
        L4 & Cooperative active sensing & agreement \\
        \hline\\
    &\hspace{2cm}         \textbf{\textsc{4. Crash avoidance}}&\\

        L1 & No cooperative sensing & default (break) \\\hline
        L2 & Cooperative sensing (e.g, RSU) short front distance & stop\\\hline
        L3 & Cooperative sensing (e.g, RSU) long front distance & slow down\\\hline
        L4 & Cooperative sensing (e.g, RSU) short front and back distance & maneuver\\\hline
        L5 & Cooperative active sensing & agreement \\\hline
    \end{tabular}
\end{table}
\section{Trustworthy Autonomous Vehicles Architecture}
\label{sec:arch}

To address the above challenges, we propose a new preliminary AV architecture that we introduce next, while we leave empirical evaluation to future work. First, we introduce the design decisions behind \acronym.
%based on the following design decisions.

\subsection{Design Rationales}

%\textbf{Supervisory Control.} 

\subsubsection{Time-aware predictive quality degradation (TPQD)}
We bridge the All-or-Nothing gap of the AI system with Time-aware predictive quality degradation (TPQD). TPQD specifies that the AI system delivery should be maximized within safety time bounds even if at degraded quality. This can leverage the tuning properties of tunable AI models like DNNs~\cite{2016branchynet-dynamic-depth-exiting-NN,huang2017multi-dynamic-depth-exiting-NN,wang2018skipnet-CNN-skipping,lin2017runtime-width-prunning-NN,cai2021dynamic-width-skipping-branches-NN}. The intuition is to reduce the likelihood of hitting the safety time bounds and consequently fail-operational because of aiming at high predictive quality and rich recognition. For instance, Savvy allows to safely identify an elephant as an obstacle object the earliest possible, rather than optimally classifying it as a elephant when it is too late; and allows to optimally identify a tunnel as is when time permits, rather than being conservative (and maybe slow-down) if otherwise classified as an obstacle object. This decision allows making use of the AI system capabilities as much as possible without violating the safety bounds. 

\subsubsection{Safety-first supervisory control}
Safety in an autonomous vehicle is paramount. To be able to make decisions without confusions, we enforce some centralized control where processes are coordinated by a \textit{safety-critical supervisory control system} (SCS). This inherits the safety-first principles of recent architectures~\cite{CLEPA-safety-first-degrad-2019,2001-AV-Arch-Simplex,shah2019safe-AV-Simplex-EGAS,deevy2019sentinel,AV-ARch-safety-use-eGas,casimiro2013karyon}, but importantly controls and monitors the time safety bounds across processes, including the AI system. The SCS can benefit from \textit{AI-based Delivery Time Estimation} (henceforth \textit{TED}) models to estimate the delivery time of AI processes. In our experience, the inference time of pre-trained DNN models is very predictable. The challenge is with launching the time of \textit{triggering events}. For this, while we maximize the use of the available sensing and actuating capabilities, we require a quick \textit{bird's eye sensing} based on which the SSC scheduling process is sparked. In this vein, we encourage introducing more sensing technologies for that very purpose.

\subsection{\acronym Architecture}

\begin{figure}[t]
\centering
\includegraphics[width=\columnwidth]{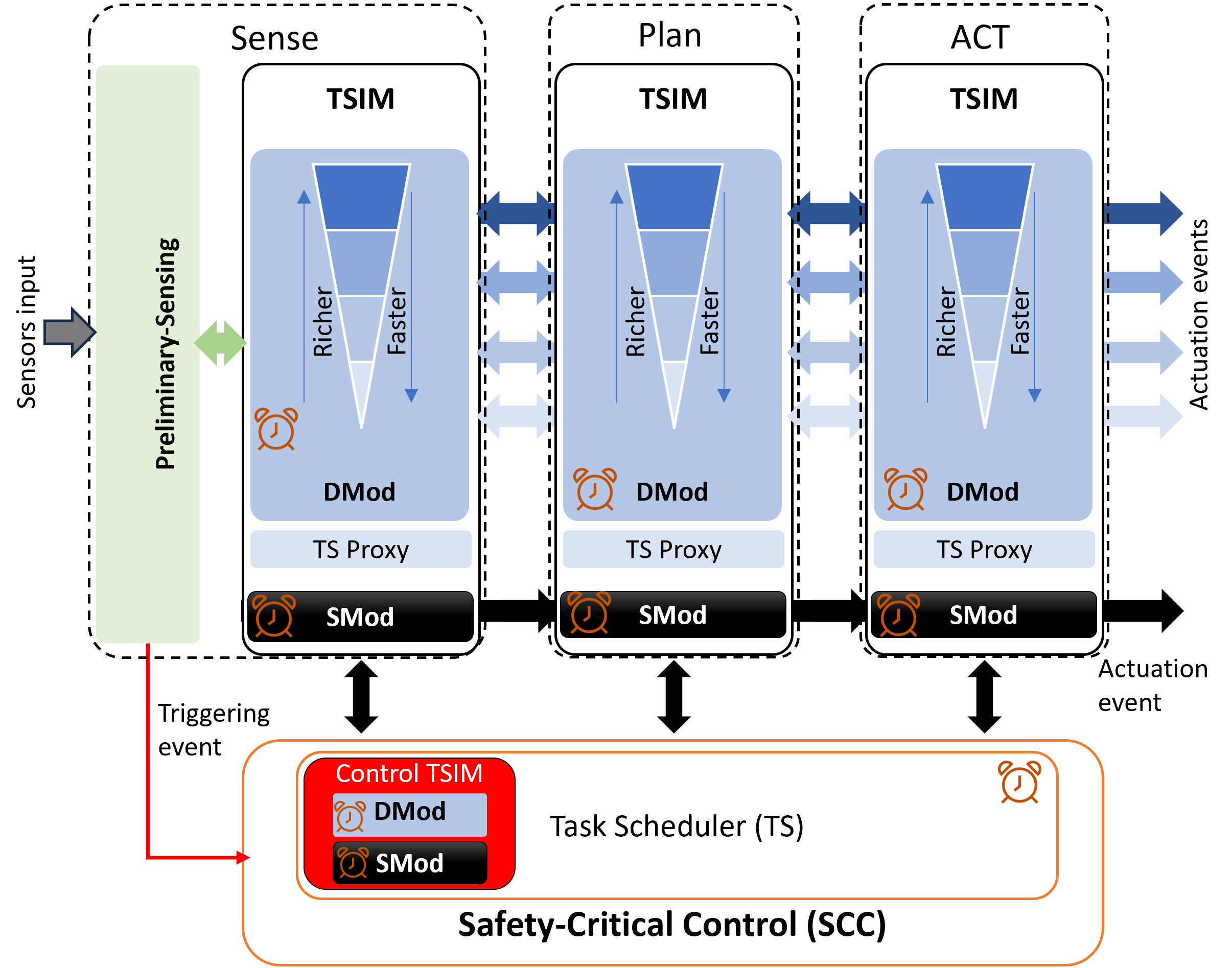}
\caption{\acronym architecture demonstrated on the Sense-Plan-Act model.}
\label{fig:arch}

\end{figure}

To implement the above design decisions, we propose the \acronym architecture, composed of two main parts: 
the \textit{Safety-Critical Control} (SCC) system and the AI-based system that can entail several \textit{Time-Sensitive Intelligent Modules} (TSIMs) depending on the perception/planning model. For instance, Fig. \ref{fig:arch} depicts the \acronym architecture with reference to the well-known Sense-Plan-Act (SPA) model~\cite{CLEPA-safety-first-degrad-2019,shah2019safe-AV-Simplex-EGAS,deevy2019sentinel}. The SCC assumes full supervisory control of the system including time scheduling of AI-based TSIMs tasks. 

The workflow starts by a sensing trigger issued by a \textit{preliminary-sensing module} that is tailored for quick bird's eye detection. This may take advantage of any sensors possible to give a heads up to the SCC, activating a new driving task process. The preliminary-sensing feeds the SCC with initial time boundaries based on which the SCC can schedule (using time prediction models) fine-grained tasks over the TSIMs, e.g., in this case the three SPA modules. Different TSIMs can interact as necessary to do more sensing, fusion, perception, planning, etc. According to the proposed time limits given to the TSIMs, the latter tune the Dynamic AI models to meet the delivery time. This may sometimes reduce the prediction quality, but should deliver useful insights, e.g., frames of animals versus animals, shadows versus liquids on road, etc. At any time, before scheduling or during running TSIM, an expired timer will immediately launch a safe-operational task controlled by the SCC (i.e., TSIM's AI never assumes control).

The SCC system plays the role of Pub/Sub broker where sensors push their readings to, and actuators take order from. Upon the receipt of input from that set of sensors, that \textit{triggers the event}, the SCC system triggers the corresponding set of actuators before the safety-critical time TTH, if better opportunistic decisions are not expected within the TTE. With this, the safety action $A$ is always guaranteed to be triggered and completed before a critical event or time TTH, beyond which the vehicle would be at risk. %The SCC may not only discard some readings or actuation, it can even kill and rejuvenate processes if the design in-variants (e.g., semantics or timing) are violated.
 The SCC system employs a time-sensitive \textit{Task Scheduler} (TS) that generates the time bounds for the driving related tasks, ensuring the safety-critical timeliness. For each driving related event, TS defines the task schedules across the different TSIMs and set the safety timers accordingly. In particular, TS defines two time bounds that represent the time interval [\textit{Time to Hazard (TTH), Time to Event (TTE)}]. The TTE, e.g., Time to Curve, Time to Overtake, etc., is the time assigned to each TSIM to deliver. TTH, on the other hand, defines the safety-critical time by which the whole driving task should complete, guaranteeing the safety. 
 %\todo[size=\small, color=green!40]{Please check it.}%

 A TSIM processes driving tasks, e.g., detection, respecting the time bounds. The TSIMs are self-contained modules that can be used in any architecture; for instance, Fig.~\ref{fig:arch} demonstrates the Sense-Plan-Act model using three TSIMs. Each TSIM is further composed of the Static submodule (SMod) and the Dynamic submodule (DMod). DMod leverages the dynamic AI models capabilities that are tuned, for instance, using Bounded AI (such as Neurosymbolic NNs~\cite{garcez2023neurosymbolic-survey,neurosymbolic-survey-2020statistical,neurosymbolic-ShieldNN-2020} and Physics Informed Neural Networks (PINN)~\cite{karniadakis2021physics-nature-PINN, PINN-quantification-abdar2021review-survey, physics-2020ppinn,physics-PINN-yang2019adversarial} models), to deliver before the TTE expires. Tuning is done via a time prediction system that learns and estimates the calibration parameters of AI models to be used in order to deliver before TTE expires. The TTE is the expected time by which all the DMods should deliver and execute in order to take advantage of the AI capabilities. In the event where the TSIM AI execution, that is DMod processing, hits the lower time bound, the SMod is triggered by firing the safety timer TTH. The TTH defines the deadline to execute all the safety executions of SMods to guarantee global fail-safe. 
 %SMod executes "spontaneous" well-defined decisions in the realm of the \textit{Organic Computing} model~\cite{muller2017organic-book,linares2018hybrid-organic-AV}, since it does not "think"; contrary to the DMod, it makes use of a special redundant channel for fast and reliable execution, as those proposed in \textit{Time-Triggered Architectures} (TTA) and \textit{Timely Computing Base} (TCB)~\cite{heiner1998time-TTA,kopetz2003time-TTA,verissimo2000timely-TCB}. 
 To estimate TTH and TTE, TS again uses a special Control TSIM whose SMod is defined at design time, and DMod is made highly accurate using real accurate formulas or more Bounded AI models. The TS distributes the TTH and TEE over all TSIMs following some policy: statically, e.g., on evenly basis, or dynamically, in a similar to DMods.

\section{Related Works}
%\todo[size=\small, color=green!40]{Rehana, plz review}%
\label{sec:related}

\subsection{Safe AV Architectures}
\textbf{Safety-first} principles have been used in early architectures of automotive safety-critical systems~\cite{2001-AV-Arch-Simplex, workgroup2013standardized-egas-arch}. The Simplex architecture includes a high-assurance and a (complex) high-performance system that run in parallel. The latter controls the system as long as safety is not violated, in which case the former can take control using a decision logic supervisory circuit. The implementation of the supervisory circuit to make a taking control decision is however complex as explained in~\cite{2014-Simplex-Complex}, due to the tradeoffs between safety and performance. Generalized concepts have been used in the E-GAS standardized architecture EGas~\cite{workgroup2013standardized-egas-arch} and~\cite{supervisor-simplex-redundancy-AV-2018}, while running multiple levels of diagnostic monitoring and redundancy. Safe-AV~\cite{shah2019safe-AV-Simplex-EGAS}, a more recent architecture, combines the prior solution in more redundancy levels while also supported ML-based AV. Similar to \acronym, these architectures make a clear separation between performance and safety planes, however without addressing ML-based systems or quality/service degradation. 

\textbf{Service level degradation}, on the other hand, has been proposed in KARYON~\cite{casimiro2013karyon} to ensure timeliness and switch to "hard-coded" safety kernel. We inspire from this work to propose degradation with ML inference, which has not been addressed in~\cite{casimiro2013karyon}. A recent monolithic architecture, Sentinel~\cite{deevy2019sentinel}, has been proposed to cover the aforementioned safety-first concepts with ML degradation. However, degradation in Sentinel is more like cross-validation, as it makes use of a combination of parallel inaccurate predictions (e.g., 60\% accuracy) to consolidate a decision. This is far from safe compared to \acronym's degradation technique that uses a degraded classification problem whose accuracy is high. For instance, Sentinel may recognize an elephant with 60\% accuracy, while \acronym recognizes it as obstructing object with 95\%. The latter is a credible accuracy to make an informed decision, although not optimal, while Sentinel's decision is highly risky.

Unfortunately, we do not discuss commercial architectures like Tesla, Waymo, Cruise, and Huawei-backed AITO M7 as these are not publicly available.

\subsection{AI/ML for AV Architectures}

The use of AI/ML in AVs is prominent in recent literature~\cite{sallab2017deep-Valeo-AV, webb2020waymo, Tesla-S-real-issues-2017autonomous, shalev2017formal-safe-AV-Mobileye, cognitive-ASAD-AV-2019-energy, fridman2019arguing-AV, desai2019soter-AI-AV, linares2018hybrid-organic-AV}. In general, most follow the \textbf{Sense-Plan-Act design}~\cite{CLEPA-safety-first-degrad-2019}, running several AI/ML architectures and models. The Sense part is focused on sensor data processing including detection, recognition, perception, and localization. It makes use of \textit{Deep Learning} for object detection and classification problems~\cite{lecun2015deep-DL, CNN-DN-2017segnet} being able learn new features without handcrafted features. In particular, \textit{Convolutional Neural Networks} (CNNs) have shown to be promising for lane and vehicle detection~\cite{huval2015empirical-driving-AV-CNN-DN}. Plan also uses ML models for prediction and planning. Prediction is essential to guess and evaluate the expected future (e.g., trajectory or behaviour) of the vehicle considering its dynamic surrounding. 
\textit{Recurrent Neural Networks}~\cite{pinheiro2014recurrent-RNN,sutton2018reinforcement-RNN-book} (RNNs) are essential to this class of problem, especially \textit{Long-Short Term Memory }(LSTM) networks~\cite{hochreiter1997long-LSTM-RNN}, used to integrate past the present information for end-to-end scene labeling systems. Recent models like Reinforcement Learning~\cite{sutton1988learning-reinforcement-RL,sutton2018reinforcement-RNN-book}, mixed with DL can
achieve human-level control in~\cite{lecun2015deep-DL,DL-AV-2020survey}, and Attention models~\cite{mnih2014recurrent-RNN-attention,xu2015show-attend-tell-planning-RL-attention} are being used to improve information filtering. This is mainly useful to focus on the relevant part to "attend" in highly dimensional data, e.g., camera images. 

Our work in progress make uses of these techniques and models with two main differences: First, we recommend those models, e.g., \textbf{Dynamic Neural Networks} (DNN), that are easily tunable at inference time to be able to leverage the full power of AI/ML despite time limits. Many of the models discussed in this section lie in the category and support dynamic features like: model deformation using depth and width adjustment~\cite{2016branchynet-dynamic-depth-exiting-NN,huang2017multi-dynamic-depth-exiting-NN,wang2018skipnet-CNN-skipping,lin2017runtime-width-prunning-NN}
%,cai2021dynamic-width-skipping-branches-NN} 
(early exiting, skipping, pruning, etc.), choosing the adequate protocol using Neural Architecture Search~\cite{zoph2016neural-NAS-survey, yu2020bignas-BigNAS} (NAS), or parameter (Weights, Space, or Channel) adjustments~\cite{woo2018cbam-attention,guo2022attention-survey,wang2017deep-attention,chen2020dynamic-attention,li2021dynamic-slimmable-NN} at inference time. Second, for more accurate recognition and prediction to improve safety, our work encourages more research and use of what we call \textbf{Bounded AI} (BAI) prediction models that include factual pre-trained or symbolic models to guide the training of the main processing models. Recent methods like Neurosymbolic NNs~\cite{garcez2023neurosymbolic-survey,neurosymbolic-survey-2020statistical,neurosymbolic-ShieldNN-2020}, Physics Informed Neural Networks (PINN)~\cite{karniadakis2021physics-nature-PINN, PINN-quantification-abdar2021review-survey, physics-2020ppinn},
%physics-PINN-yang2019adversarial}, 
Constrained or Conservative PINNs~\cite{contrained-CPINN-zhu2019physics,CPINN-jagtap2020conservative}, Finite Basis PINN~\cite{physics-informed-FBPINN-2021-finite}, Variational PINN~\cite{physics-VPINN-kharazmi2021hp} are believed to have close to 100\% accuracy in some contexts.

\section{Conclusion}
\label{sec:conc}
%\todo[size=\small, color=green!40]{Rehana, plz write this}%
AI is proving to be widely useful especially for non-critical applications. AVs are however more challenging being safety-critical and often time-critical. Reality shows that AI can be useful to improve safe driving compared to humans; however, our analysis shows that AI is not reliable per se to take control of the AV. In particular, due to time-criticality of AV tasks, the architects are either in the conservative camp, and tend to refuge to fail-operational mode often, or in the optimistic camp where performance and user convenience are prioritized at the cost of safety. This paper presents a new AV architecture \acronym following a new approach, we call Time-aware predictive quality degradation (TPQD), to combine the two advantages without violating the other. \acronym leverages the Dynamic NN properties through tuning them at inference time given the available safety time boundaries. This leads to a trustworthy AV where even for limited time windows, the AI power is being exploited, thus avoiding the \textit{all-or-nothing dilemma}. We are currently implementing the architecture and exploring the DNN models to evaluate empirically.

%%%%%%%%%%%%%%%%%%%

\bibliography{IEEEabrv,ref.bib}
\bibliographystyle{IEEEtranS}

%%%%%%%%%%%%%%%%%%%

\end{document}